\documentclass{article}
\usepackage{arxiv}

\usepackage[utf8]{inputenc} 
\usepackage[T1]{fontenc}    
\usepackage{hyperref}       
\usepackage{url}            
\usepackage{booktabs}       
\usepackage{amsfonts}       
\usepackage{nicefrac}       
\usepackage{microtype}      
\usepackage{lipsum}		
\usepackage{graphicx}
\usepackage{natbib}
\usepackage{amsmath}
\usepackage[colorinlistoftodos]{todonotes}
\usepackage{doi}

\begin{document}
\pagenumbering{arabic}
%
\title{L2Explorer: A Lifelong Reinforcement Learning Assessment Environment}
\author{ \\ {\bf Erik C. Johnson, Eric Q. Nguyen, Blake Schreurs, Chigozie S. Ewulum, Chace Ashcraft, Neil M. Fendley,} \\ {\bf Megan M. Baker, Alexander New, Gautam K. Vallabha} \\
Johns Hopkins University Applied Physics Laboratory\\
11100 Johns Hopkins Rd.\\
Laurel, MD 21045\\
}
\maketitle
\begin{abstract}
\begin{quote}
Despite groundbreaking progress in reinforcement learning for robotics, gameplay, and other complex domains, major challenges remain in applying reinforcement learning to the evolving, open-world problems often found in critical application spaces. Reinforcement learning solutions tend to generalize poorly when exposed to new tasks outside of the data distribution they are trained on, prompting an interest in continual learning algorithms. In tandem with research on continual learning algorithms, there is a need for challenge environments, carefully designed experiments, and metrics to assess research progress. We address the latter need by introducing a framework for continual reinforcement-learning development and assessment using Lifelong Learning Explorer (L2Explorer), a new, Unity-based, first-person 3D exploration environment that can be continuously reconfigured to generate a range of tasks and task variants structured into complex and evolving evaluation curricula. In contrast to procedurally generated worlds with randomized components, we have developed a systematic approach to defining curricula in response to controlled changes with accompanying metrics to assess transfer, performance recovery, and data efficiency. Taken together, the L2Explorer environment and evaluation approach provides a framework for developing future evaluation methodologies in open-world settings and rigorously evaluating approaches to lifelong learning. 
\end{quote}
\end{abstract}

\noindent
In recent years, Deep Reinforcement Learning (DRL) approaches have begun to deliver powerful results for a variety of compelling domains, including games such as Chess, Go, and Shogi~\cite{silver2018general};  Atari video games~\cite{mnih2013playing}; more complex strategy video games~\cite{berner2019dota,vinyals2019grandmaster}; and dexterous robotic manipulation~\cite{rajeswaran2017learning}. Despite the groundbreaking success in training autonomous agents, resulting policies tend to be very brittle and generalize poorly~\cite{chan2019measuring}. When presented with a new task or a task variant, DRL approaches are susceptible to a performance drop~\cite{zhang2018study,kirk2021survey} due to the catastrophic forgetting problem~\cite{French1999CF,McCloskey1989CF}, which may not be overcome by domain randomization strategies alone. As the field moves from environments which are fixed to evolving, open-world scenarios, current DRL approaches will be insufficient. 

\begin{figure}
    \centering
    \includegraphics[width=0.8\textwidth]{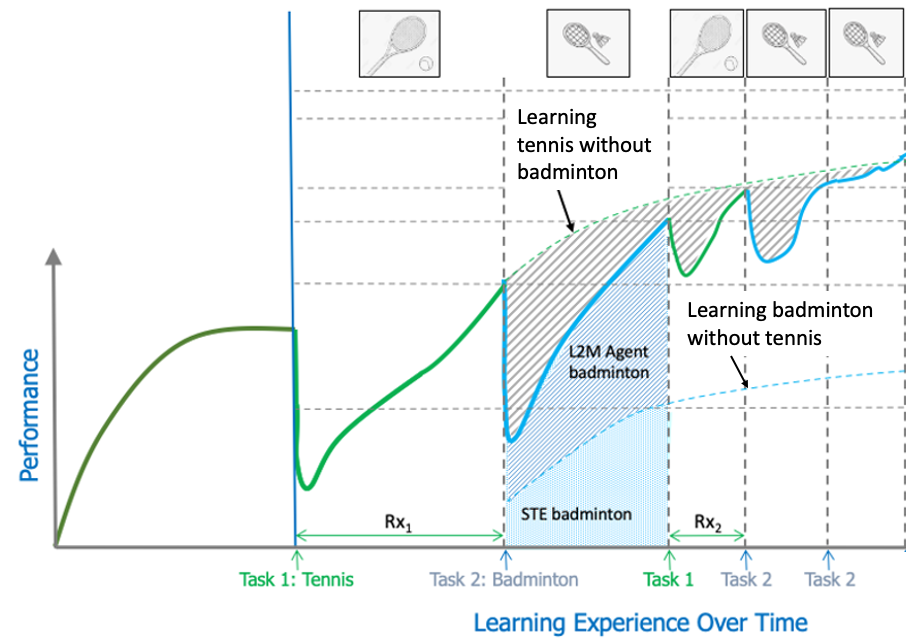} 
    \caption{Key considerations when assessing lifelong learning agents (a Lifelong Learning Machine, L2M) with open-world environments. An agent's lifetime is broken into blocks of tasks (e.g., playing an individual sport) for training and evaluation. Blocks generate metrics for assessing performance. Our approach utilizes a highly flexible environment to generate curricula (lifetimes), tasks, and task variants. A multi-dimensional approach to metrics, going beyond raw performance, is required to capture the nuances such as transfer and performance recovery. Controlled sequences of tasks can elucidate particular strengths and weaknesses in learners. Figure adapted from~\cite{New2022lifelong}.}
    \label{fig:metrics}
\end{figure}

This performance gap has led to an interest in \textit{Continual Learning}, which seeks to design algorithms to learn over sequences of tasks. In the related, but broader, concept of \textit{Lifelong Learning}~\cite{chenliu2018lifelong}, an agent learns over a lifetime of experiences (see Fig. \ref{fig:metrics}) in an evolving environment (for purposes of this paper, however, we treat \textit{continual learning} as synonymous with \textit{lifelong learning} as our approach is applicable to both concepts). Much recent work has been on supervised classification under distribution shifts~\cite{song2020critical} and learning a sequence of tasks \cite{parisi2019continual,hsu2018re}. Continual RL~\cite{khetarpal2020towards} seeks to create agents which can maintain performance in the face of nonstationary distributions.

A key issue in developing Continual RL algorithms is how to assess performance in a rigorous and informative way. There are existing attempts to address this with procedurally generated open worlds~\cite{risi2020increasing}, which generate tasks using 
randomly parameterized environments. A recent software framework has begun integrating metrics, baselines, and environments for Continual RL~\cite{powers2021cora}. We argue that successful Continual RL assessment requires both sufficiently complex environments and highly reconfigurable tasks, as well as carefully structured experiments and metrics. We must also consider a multi-dimensional approach to assessment (in terms of raw performance, generalization, task transfer, and data efficiency). 

To help address this need, we have developed Lifelong Learning Explorer (L2Explorer), a first-person-view (FPV), highly configurable Unity\textsuperscript{TM} environment\footnote[1]{\url{https://unity.com/}}. The environment allows for procedural generation through Python code for open-ended task and task variant definitions. Moreover, the environment is set up for testing lifelong learning curricula with a set of lifelong learning metrics. While the Unity environment itself is designed for Continual RL assessment, this work also outlines an approach for taking complex, reconfigurable open-world environments and creating rigorous evaluation for lifelong learning algorithms.

\section{Existing Environments for Continual RL}

Many environments have been used to test DRL approaches, including game environments. Extensive work, including transfer learning studies, have been conducted with Atari games~\cite{mnih2013playing}. This has been generalized into a meta-learning framework that allows sampling of Atari-like games~\cite{staley2021meta}. Complex strategy games have also been leveraged in this context, including Starcraft 2\footnote[2]{\url{deepmind.com/research/open-source/pysc2-starcraft-ii-learning-environment}}, team-based play in Dota 2 \cite{berner2019dota}, and rogue-likes such as Dungeon Crawl Stone Soup~\cite{dannenhauer2019dungeon}. While impressive, many games can be limited for Continual RL testing due to the lack of full configurablility. 

Simulators of real-world systems, such as autonomous vehicles, have also been extensively used. Two notable examples are the CARLA simulator~\cite{Dosovitskiy17CARLA} for autonomous driving and the AirSim package for unmanned aerial vehicles~\cite{shah2018airsim}. While these include photorealistic images and realistic physics, the computational complexity of the environments coupled with the design time required to produce tasks and task variants can limit their applicability to Continual RL testing. 

There are also many FPV environments, which combine some of the desirable features of game environments and physical simulators. These are valuable for continual and lifelong learning testing because they allow natural inclusion of partial observations, multiple observation modes, and proxies of real world tasks. These strengths come at the downside of reduced task complexity and reduced fidelity. Examples include the use of the Unity Environment through the ML-Agents package~\cite{juliani2018unity} and the procedurally generated Obstacle Tower~\cite{juliani2019obstacle} environment. The FPV game DOOM forms the basis of the Vizdoom FPV environment~\cite{kempka2016vizdoom}. Also related are the real-world home and robot environment simulators such as AI2Thor~\cite{kolve2017ai2}, which also introduce semantic object relationships into FPV environments. 

In order to introduce the flexibility required for open-world learning scenarios, increasing emphasis has been placed on procedurally generated environments. These include the open-world game environments of XLand~\cite{team2021open} and MiniHack~\cite{samvelyan2021minihack}. These represent major steps forward, but the lack of controllable parameters limits the precision of testing that can be done in these environments~\cite{kirk2021survey}. 

Our environment and assessment framework, L2Explorer, seeks to combine the strengths of different types of environments. The use of Unity allows for a visually complex world, and ML-Agents exposes hooks that allow for principled procedural generation using flexible Python code. The environment has lower computational complexity than simulators of autonomous cars or aerial vehicles. Finally, the critical component is linking this procedurally generated environment to careful experimental design and metrics, similar to CoRA~\cite{powers2021cora}. We aim to create an approach that will also generalize and improve Continual RL assessment utilizing other open-world environments. 

\section{L2Explorer Design Approach}
Similar to previous efforts to define key requirements for lifelong learning evaluations~\cite{farquhar2018towards}, we developed key criteria for our framework which allow for rigorous assessment of lifelong learning with an open-ended, procedurally generated environment. We believe the key criteria of our framework to be: 
\begin{itemize}
\item Flexible description of multiple tasks: A programmatic API to specify new tasks with a common specification
\item Flexible control over task variants: Selection of key variables that can be modified within a task definition
\item Notions of task relationships/similarity: While there is no agreed upon approach to measure generic task similarity for RL tasks, some notion of task similarity is required to appropriately structure evaluation
\item Control over degree of similarity: The environment needs to be able to control the degree of similarity between configurations to allow for abrupt and gradual transitions
\item Parametric and random variation: The tasks require the notion of parameters used to deterministically create variants, as well as intrinsic parameters that can be randomly sampled, similar to the previous work~\cite{kirk2021survey}
\item Targeted testing curricula: A set of designed curricula and tests, along with integrated metrics and baselines
\end{itemize}
Taken together, these factors address some of the limitations that come with testing with open-world, procedurally generated environments.

\subsection{Continual RL Assessment}
In L2Explorer, we build on the standard formalism for RL agents using Partially Observable Markov Decision Processes (POMDPs). These consist of the tuple $M=(S,A,O,R,T,\phi,p)$, where $S$ is the state space, $A$ is the space of actions available to an agent, $R: S \times A \times S \to \mathbb{R}$ is the reward function, $T(s'|s,a)$ is the state transition function, and $p$ is the distribution of initial states. The system is partially observable as the state is not directly available to the agent, which instead gets an observation from the observation space $O$ generated by the emission function $\phi: S \to O$. The goal of the RL agent is to learn a policy $\pi(a|s)$.
\begin{equation*}
    \pi^* = \text{argmax}_{\pi \in \Pi} \mathbb{E}_{s \sim p(s_0)}[R(s)]
\end{equation*}
where $R(s)$, which is a real value, is the total expected reward gained by the state. 

We consider nonstationarities in the underlying POMDP for both tasks and tasks variants. Of critical consideration are $S$, $A$, $T(s'|s,a)$, $R$, $O$, and $\phi$. In continual reinforcement learning, these are allowed to vary as a function of the number of episodes to create a nonstationary POMDP.

To expose agents to nonstationarity in a meaningful way, controlled variation in environment is required. It is rarely practical to fully quantify distributions in high dimensional POMDPs, so instead we seek useful and practical surrogates for controlled manipulation of nonstationary POMDPs. We propose that each curricula should have some desired testing hypothesis relating to changing distributions in the underlying POMDP. Different tasks and task variants can represent particular distributions in the POMDP tuple, and altering parameters results in a shift in these distributions.

Achieving this requires some notion of task and variant similarity. We propose exploiting the parameters of procedural generation to create heuristics of similarity. For example, if two task differ only by the probability distribution used to place objects, a distance metric applied to the tasks' distributions could be a proxy measure for the similarity. A key caveat, however, is that heuristic notions of similarity may not be inherently related to agent transfer. What seems intuitively similar to a human may not correspond to positive transfer between tasks for an RL algorithm, and tasks that may be similar for one algorithm may not be for another \cite{carroll2005task}. Research into measures of similarity, transfer, and generalization is ongoing~\cite{barreto2016successor,ma2020universal} and beyond the scope of this work.

L2Explorer variants are designed by selecting a subset of procedural generation parameters that must be fixed, and a subset that may be fixed or allowed to be sampled from a distribution. Differences in the parameter values may be passed into the appropriate similarity heuristics. Tasks are assembled into curricula with particular experimental goals, such as determining sensitivity to changes in reward space, the observation emission function, and so forth. This overall approach will allow for meaningful experimental design and overcome limitations of completely randomly generated environments~\cite{song2020critical,kirk2021survey}. 

\subsection{Curriculum Design Considerations}
We utilize existing a lifelong learning evaluation framework for curriculum design, as specified in~\cite{New2022lifelong} and released open-source\footnote[1]{ \url{https://github.com/darpa-l2m/l2metrics}, \url{https://github.com/darpa-l2m/l2logger}}. This can be seen in Fig.~\ref{fig:metrics}, where an agent is pretrained and then deployed on a sequence of tasks. A lifelong learner should be able to take such a fixed curriculum and demonstrate learning over deployment. This framework allows for pre-deployment training or parameter selection before learning. The agent is then subjected to a series of units of experience (e.g., rounds in a game) that can be grouped into longer blocks. We assume blocks can incorporate drift in the underlying distributions of the POMDP. Periodically, the agent can be frozen and have its capabilities tested in a separate evaluation block. Individual environment metrics can be specified, then aggregated into environment-agnostic lifelong learning metrics (detailed below). In L2Explorer, similarity heuristics can be utilized to order tasks and task variants into sequences of learning and evaluation blocks. 


\subsection{Metrics}
\label{sec:metrics}
L2Explorer utilizes an existing set of lifelong learning metrics that characterizes the nuances of the performance curves such as the one in Fig.~\ref{fig:metrics}. Full definitions can be found in~\cite{New2022lifelong}, with open-source implementations available. Task-specific performance measures (e.g., total reward, time to completion) generated during learning and evaluation blocks enable computation of:
\begin{itemize}
    \item Performance Maintenance: A measure of catastrophic forgetting for a given task
    \item Forward Transfer: A measure of learning improvement for a task, given learning experiences on a different task
    \item Backward Transfer: A measure of improvement on a previously learned task given learning experiences on a new task
    \item Performance Relative to Single-Task Expert: Comparison of lifelong learning agent to an expert agent trained only on the task in question (e.g., the dashed lines in Fig.~\ref{fig:metrics}).
    \item Sample Efficiency: A measure of the experience required for a lifelong learning agent to reach maximal performance relative to a single task expert
\end{itemize}

Together, the metrics, curriculum design, and design philosophy inform the development of the L2Explorer environment. Lessons from this approach can be generalized to assessment with future open world exploration environments.

 \begin{figure*}
    \centering
    \includegraphics[width=0.8\textwidth]{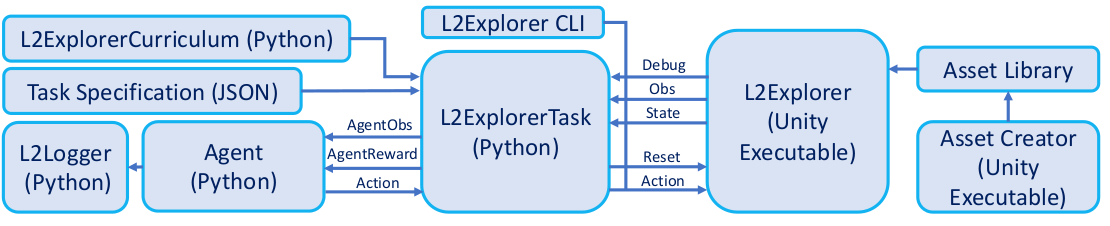} 
    \caption{Overview of L2Explorer Software system, consisting of a suite of Python tools which interact with a custom Unity environment. The Python suite allows for execution of testing curricula, integration of standard agent code, and rapid reconfiguration of the unity environment through the reset channel. Logging and metrics computations are integrated. A custom asset creator allows new Unity models to be incorporated into the framework to maximize extensibility.}
    \label{fig:software}
\end{figure*}

\section{L2Explorer}
L2Explorer is a software framework and testing environment which implements this overall design vision, including integration with metrics and testing curricula. The project builds on Unity ML-Agents~      \cite{juliani2018unity}, which exposes key parameters to create a reconfigurable, 3D, FPV reinforcement-learning environment. Our custom Unity environment enables exploration and interaction with a world specified through a JSON object format. Additional side channels allow communication of observations, actions, and debugging information. The Python code specifies the world JSON objects, allows for creating testing curricula, integrates metrics, and provides a Gym-compatible interface~\cite{brockman2016gym}. An overview of the software can be seen in Fig.~\ref{fig:software}, and is intended to be released open-source. The package will provide testing curricula and baselines to assist development, and will allow users to fully customize tasks, task variants, curricula, and even import custom 3D models into the environment. 

\subsection{Unity Executable}

\begin{figure}[ht!]
    \centering
    \includegraphics[width=0.35\textwidth]{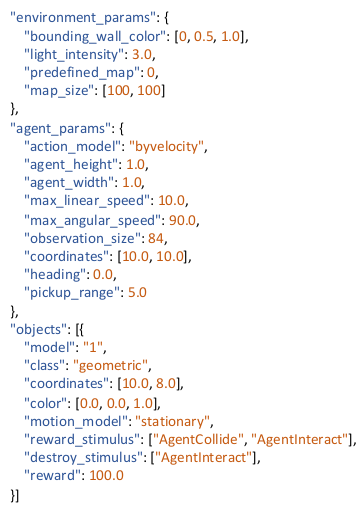} 
    \caption{Example subset of a reset JSON object. This shows the available variables to procedurally generate variants of the environment, agent, and objects. Each object can have independently specified coordinates and interaction models, and objects can be created dynamically within an episode. On each reset call, a new JSON can be loaded into the environment. }
    \label{fig:reset}
\end{figure}

The core environment of L2Explorer is a custom Unity app which instantiates a simulated world in response to a JSON specification communicated through a Unity ML-Agents side channel. This is activated when the reset function is called, allowing each learning experience to take place in a custom specified world. A partial example of a reset JSON can be seen in Fig.~\ref{fig:reset}. The format allows for specification of environment specific parameters, including lighting conditions, backgrounds, agent specific parameters, and object specific parameters. Pre-caching 3D models enables rapid world construction on reset.

The Unity environment exposes three custom communication channels seen in Fig.~\ref{fig:software}. These allow the reset functionality, querying the current state of the world (returned in the specified world JSON format), and sending debugging information. Standard channels allow for the communication of actions (continuous or discrete) and observations (state vectors and visual observations). Visual observations include RGB images, grayscale depth images, and semantic segmentation maps (segmented by the object ``class''). 

The agent model is, by default, a kinematic model controlled by a linear and angular velocity term. Several agent-object interaction models are specified. These can require an agent to be in a nearby zone, require contact with the object, or require the agent to take an ``interact'' action. Interaction can result in a positive or negative reward, and can result in the destruction of the object. There is dynamic respawning through the side channels, allowing new objects to be specified during runtime. For certain task designs, this can be key for learning with fixed episode step length. 

A separate unity executable is provided to allow users to prepare and save custom 3D models in an L2Explorer compatible format. These assets, once prepared, can be loaded in during a reset action. This can allow for continued extensibility of the environment. 

\subsection{Python Wrappers}
In L2Explorer, Python wrappers allow for task specification and agent development. The primary class, L2ExplorerTask, is a wrapper for communicating with the Unity executable. This class processes the raw observations to provide a reward and observation to the agent. This allows for definitions of alternative rewards functions in Python, such as soft rewards, as well as any data filtering required for a task. 

The Unity environment is configured during a reset function, utilizing the JSON specification (Fig.~\ref{fig:reset}). This enables the specification of curricula in Python, as well as the specification of procedural generation. Curricula are specified through task-specific classes and a task-specific initial JSON file. Learning and Evaluation Blocks can then be described as above, along with variants of the task. The base JSON file for the task is passed to the task class, and key-value pairs are modified, if required, to create a variant. Modifications are drawn from discrete or continuous probability density functions to create appropriate randomization. The Gym-compatible environment class allows for integration of standard baselines and custom algorithms. The agent observation space, $O$, is a tensor of number of pixels by number of pixels by number of channels in the selected modalities, and optionally state vector of size $N$. The action space $a$ is a 2-dimensional continuous valued vector (linear and angular velocity), which can optionally be discretized. A third dimension can be added for the binary ``interact'' action.   

In addition to the core functionality, integration with a Python logging package and Python metrics package for lifelong learning is provided to enable the multi-dimensional metrics evaluation. A command line interface (CLI) is also provided for directly interfacing with and debugging the environment. 

\section{Example Use Cases}
L2Explorer is a highly configurable and extensible environment that can provide insight into the performance of Continual RL algorithms. Towards this end, we present some example tasks and curricula that will be refined into standardized benchmarking experiments.

\begin{figure*}
    \centering
    \includegraphics[width=\textwidth]{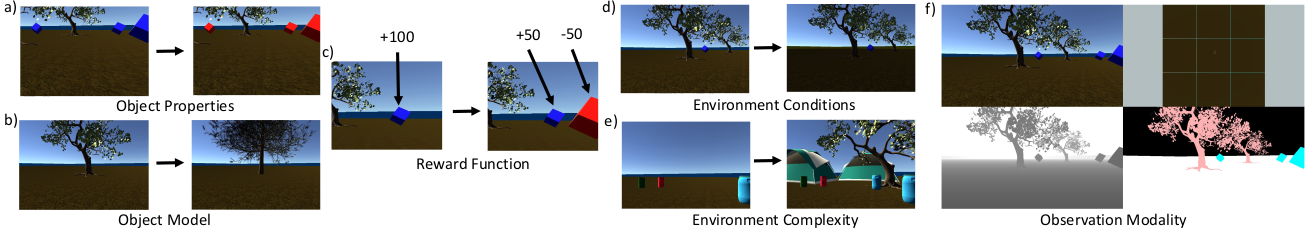} 
    \caption{Variants which can be produced using the standardized JSON format through the reset function. Panel a) shows how object properties can be manipulated, in this case turning blue targets into red ones. Panel b) shows the replacement of an object of the same class with a different model, in this case swapping two trees. Panel c) shows how the reward structure can be modulated with one reward object of value $100$ becoming two reward objects with values $50$ and $-50$. Panel d) shows the manipulation of environmental conditions through changes in lighting. Panel e) shows how scene complexity can be varied through placement of additional objects from different classes. Finally, Panel f) demonstrates how different observation modalities including RGB, depth, and a semantic segmentation can be supplied to an agent.}
    \label{fig:variants}
\end{figure*}

\subsection{Configuration of L2Explorer }
Fig.~\ref{fig:variants} shows some aspects of the world that can be reconfigured. These relate (indirectly) to the underlying POMDP, including $O$, $\phi$, and $S$, by altering the object properties, object models, environmental complexity, environmental conditions, and observation modality. The reward $R$ and state transition function $T$ are more directly manipulated through modifying the object interaction model, object reward, and agent models. The extensible L2ExplorerTask class allows for further manipulations such as soft reward functions and observation state vectors. 

\begin{figure}
    \centering
    \includegraphics[width=0.45\textwidth]{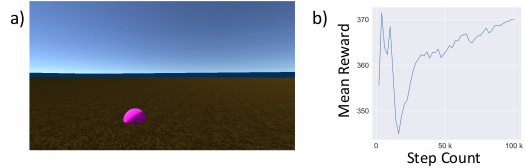} 
    \caption{Example task demonstrating integration of a Stable Baselines PPO agent for learning a simple object finding task. Panel a) shows the visual input to the agent and Panel b) an example learning curve. Standard Python agents using the established Gym API can be rapidly tested in this framework.}
    \label{fig:learning}
\end{figure}

\subsection{Tasks}
For each task a JSON is specified from which variants can be created and randomization can be done. Each task has a set of parameters which define the core task, parameters to change to produce static variants, and parameters that can be randomized within a variant. Fig.~\ref{fig:learning} shows an example of performance curves for a Stable Baselines Proximal Policy Optimization\footnote[1]{\url{https://github.com/DLR-RM/stable-baselines3}}. Developers of novel algorithms will be able to exploit this interface to baseline directly against single task experts. While L2Explorer is designed to be extensible, several initial tasks have been developed: 

\begin{itemize}
    \item Find Objects: The agent interacts with objects to receive a positive reward. Environmental conditions, object properties, and agent properties can be randomized. Variant tasks include observation changes of the target model, state space changes of the map layout, and agent-interaction model changes
    \item Get to Goal: The agent receives positive reward for navigating to a target position. Goal positions, map layouts, and agent parameters can be randomized. Variants include changes in visual landmarks, soft reward functions, and the addition of negative reward objects
    \item Select Object: The agent receives positive reward for selecting the correct object, negative reward for other objects. Reward probabilities can be manipulated as in a $K$-armed bandit problem. Variants include changes in the reward probability distribution and object models
    \item Moving Object: Similar to Find Objects, with a object motion model. Variants allow for manipulation of the action space and state transition space through changes in the agent and object motion models
    \item Scavenger Hunt: The agent can obtain a series of rewards by interacting with the right sequence of objects. Variants include changes in visual landmarks, changes in the reward distribution, and changes in the object models
\end{itemize}

\begin{figure*}
    \centering
    \includegraphics[width=0.8\textwidth]{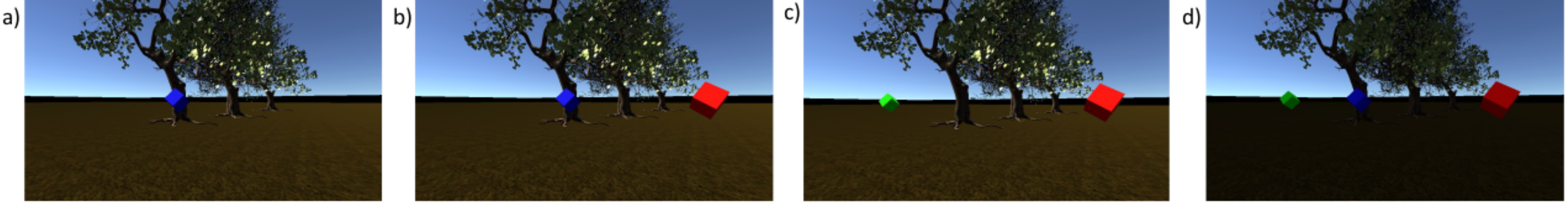} 
    \caption{Example curricula showing variations in observation and reward models to realize the evaluation framework (Fig.~\ref{fig:metrics}). Between each variant, the reward structure and observation structure change, enabling analysis of performance and manipulating the underlying POMDP. In Panel a) there is a positive reward associated with capturing the blue object. In the variant in Panel b) a red object object is introduced which returns a negative reward when captured. In the variant in Panel c) a green object returns a positive reward. Finally, in Panel d) the variants in Panel b) and Panel c) are mixed with different environmental conditions.}
    \label{fig:curriculum}
\end{figure*}

\subsection{Assessment Curricula}
Tasks and task variants can be assembled into learning and evaluation blocks such as in Fig.~\ref{fig:curriculum}. In this example changes in tasks can, through assumed heuristics such as a norm of the object RGB color difference, approximate changes in the observation space and reward space. The magnitude of parameter changes can also be used as a calculable surrogate for similarity, although we reemphasize that such heuristics do not necessarily indicate successful transfer learning. 

Preliminary curricula are being developed for several key sources of variability impacting Continual RL agents. First is manipulation of the reward structure. Distributions of rewards and their associated objects can be shifted in gradual or continuous ways. Changes in the state space and state transition space can be probed through manipulations of the maps, agent and object motion models, and interaction models. Changes to the observation emission function can be probed through changes in environmental factors, observation modalities, and object models. Action space changes can relate to the agent interaction model. The metrics will allow the assessment of changes of different magnitudes.

Similarity measures are critical to the construction of these curricula. We propose utilizing changes in the parameters of the reset JSON to provide a usable, if limited, approximation of similarity. Changes to individual object parameters between tasks can be measured with chosen distance metrics to approximate differences in similarity between the tasks. Further, we may provide a mapping from non-numeric values, such as a color specification of ``green'', to numeric ones, such as the array \texttt{(0, 128, 0)}, which can then be used with a distance measure to create another measure of similarity between tasks. Object coordinates can also be discretized into occupancy maps to allow for distance computations. While this approach is clearly limited, it is simple and useful for organizing curricula for algorithm assessment. 

\section{Conclusions}
We have presented the design approach and implementation of L2Explorer, a highly configurable, Unity-based environment for testing continual and lifelong learning systems. In addition to the software system itself, we have argued for a framework to approach lifelong learning assessment in open-ended worlds. A key consideration is the integration of appropriate metrics, experiment design, and curriculum design into the framework. 

To fully demonstrate the power of this approach, integration of single task baselines and Continual RL baselines~\cite{khetarpal2020towards}, including replay based approaches such as CLEAR~\cite{rolnick2018experience}, is needed. In combination with baseline development, the project aims to provide a set of canonical experiments for RL assessment in a FPV environment. Moreover, we anticipate the design framework can be reused. Without progress in assessment frameworks, it will be a challenge to push the frontier of reinforcement learning in open-world environments. 

\section{Code Availability}
The code for the L2Explorer framework will be made open source at \url{https://github.com/lifelong-learning-systems/l2explorer}.

\section{ Acknowledgments}
Development of L2Explorer was funded by the DARPA Lifelong Learning Machines (L2M) Program. The authors would like to thank Olivia Lyons, Ji Pak, Sasha Smith, Brianna Raphino, and Sydney Floryanzia for their help in the early development of L2Explorer, and Angel Yanguas-Gil for valuable feedback. The views, opinions, and/or findings expressed are those of the authors and should not be interpreted as representing the official views or policies of the Department of Defense or the U.S. Government. 
\bibliographystyle{unsrtnat}
\bibliography{main}

\end{document}